\documentclass[11pt]{article} 
\usepackage[letterpaper,margin=1in]{geometry}
\usepackage[round,authoryear]{natbib}
\usepackage{lmodern,microtype}

\usepackage{amsmath,amsfonts,bm}

\def\eqref#1{equation~\ref{#1}}

\def\1{\bm{1}}

\DeclareMathAlphabet{\mathsfit}{\encodingdefault}{\sfdefault}{m}{sl}
\SetMathAlphabet{\mathsfit}{bold}{\encodingdefault}{\sfdefault}{bx}{n}

\usepackage{hyperref}
\usepackage{url}
\usepackage{booktabs}
\usepackage{graphicx}
\usepackage[table]{xcolor}
\usepackage{caption} 

\usepackage{booktabs}
\usepackage{multirow}
\usepackage{graphicx}
\usepackage{xcolor}
\usepackage{colortbl}
\usepackage{array} 
\usepackage{makecell} 
\usepackage{amsmath}
\usepackage{amssymb}
\usepackage{titlesec} 
\usepackage{hyperref}
\hypersetup{
colorlinks=true,
linkcolor=black,
filecolor=black,
urlcolor=blue,
citecolor=blue
}

\newlength{\origabovedisplayskip}
\newlength{\origbelowdisplayskip}
\newlength{\origabovedisplayshortskip}
\newlength{\origbelowdisplayshortskip}
\AtBeginDocument{
  \setlength{\origabovedisplayskip}{\abovedisplayskip}
  \setlength{\origbelowdisplayskip}{\belowdisplayskip}
  \setlength{\origabovedisplayshortskip}{\abovedisplayshortskip}
  \setlength{\origbelowdisplayshortskip}{\belowdisplayshortskip}
}
\newcommand{\TightDisplayMath}{
  \setlength{\abovedisplayskip}{4pt}
  \setlength{\belowdisplayskip}{4pt}
  \setlength{\abovedisplayshortskip}{2pt}
  \setlength{\belowdisplayshortskip}{2pt}
}
\newcommand{\NormalDisplayMath}{
  \setlength{\abovedisplayskip}{\origabovedisplayskip}
  \setlength{\belowdisplayskip}{\origbelowdisplayskip}
  \setlength{\abovedisplayshortskip}{\origabovedisplayshortskip}
  \setlength{\belowdisplayshortskip}{\origbelowdisplayshortskip}
}

\titlespacing*{\section}{0pt}{1.2ex plus 0.3ex minus 0.2ex}{0.8ex plus 0.2ex}
\titlespacing*{\subsection}{0pt}{1.0ex plus 0.2ex minus 0.2ex}{0.6ex plus 0.2ex}

\newcommand{\method}{FGMO}

\title{Balancing Multimodal Learning via Functional Progress}

\author{Zhongjing Gu \quad Fengqiang Wan \quad Yiming Cui\\Yufa Feng \quad Yang Yang}
\date{}

\renewcommand{\eqref}[1]{\textup{(\ref{#1})}}

\begin{document}

\maketitle

\begin{abstract}
Multimodal learning often suffers from modality imbalance, where the joint optimization process is dominated by a single modality. Existing methods typically estimate modality imbalance from score disparities derived from prediction uncertainty or optimization statistics. However, due to distinct prediction uncertainty and learning dynamics across modalities, direct comparison of such scores may misinterpret intrinsic modality differences as progress gaps, leading to biased imbalance estimation. In this paper, we propose \textbf{F}unction-Space \textbf{G}uided \textbf{M}ultimodal \textbf{O}ptimization \textbf{(FGMO)}, which leverages a function-space progress signal to assess modality-wise optimization progress and coordinate optimization across modalities to alleviate modality imbalance. Specifically, we introduce Functional Progress Estimation (FPE) to measure each modality's update-induced function-space response and calibrate it against a loss-aligned unimodal reference, producing a comparable progress signal. Based on this signal, Functional Response Control (FRC) redistributes modality-level function-space budgets and realizes the target responses through tensor-wise learning-rate adjustment. Theoretical analysis establishes a one-step target-contraction property
of FRC under bounded controller-state mismatch,
and extensive experiments demonstrate the effectiveness of FGMO
across multiple multimodal benchmarks.
\end{abstract}

\section{Introduction}
Multimodal learning~\citep{baltrusaitis2019multimodal} aims to integrate information from multiple modalities, such as vision, language, and audio. Yet, incorporating more modalities does not always lead to better performance~\citep{wang2020gblend}. Previous studies have shown that different modalities often learn and converge at different rates during joint training~\citep{yang2024lfm}. As a result, the easier-to-learn modality may dominate the training process, while the other modalities remain under-optimized~\citep{hua2024reconboost,huang2025inforeg}. This phenomenon is commonly referred to as modality imbalance~\citep{yang2024lfm,jiang2025aug}.

To alleviate modality imbalance, existing methods typically estimate modality-wise imbalance scores from prediction uncertainty or optimization statistics and adjust the optimization accordingly~\citep{peng2022ogm,fan2023pmr,wei2025arl,gao2026rgm}. Prediction-based methods derive such scores from signals such as confidence or prediction variance~\citep{peng2022ogm,fan2023pmr,wei2025arl}, while optimization-based methods use statistics such as gradient or update magnitudes~\citep{wei2024mmpareto,gao2026rgm}. Typically, these scores are compared directly across modalities to determine which modality is under-optimized. However, different modalities naturally exhibit distinct prediction uncertainty and learning dynamics, making cross-modal score gaps poorly calibrated to the actual rebalancing demand. Consequently, the magnitude of a score disparity may not accurately indicate which modality is relatively under-optimized or how strongly it should be adjusted, potentially leading to inappropriate rebalancing.

To address this limitation, we assess each modality's optimization during joint training
relative to its own unimodal training, rather than directly comparing absolute scores across modalities.
Specifically, we assess modality optimization from a function-space perspective~\citep{milsom2025function}, which directly captures the model-output change induced by the current update while being independent of heterogeneous parameterizations. 
We then normalize each modality's current functional response by its loss-aligned unimodal reference to obtain a function-space progress signal termed Relative Functional Progress (RFP), which accounts for modality-specific learning dynamics and provides a more comparable signal for modality rebalancing.

\begin{figure*}[t]
\centering
\includegraphics[width=\linewidth]{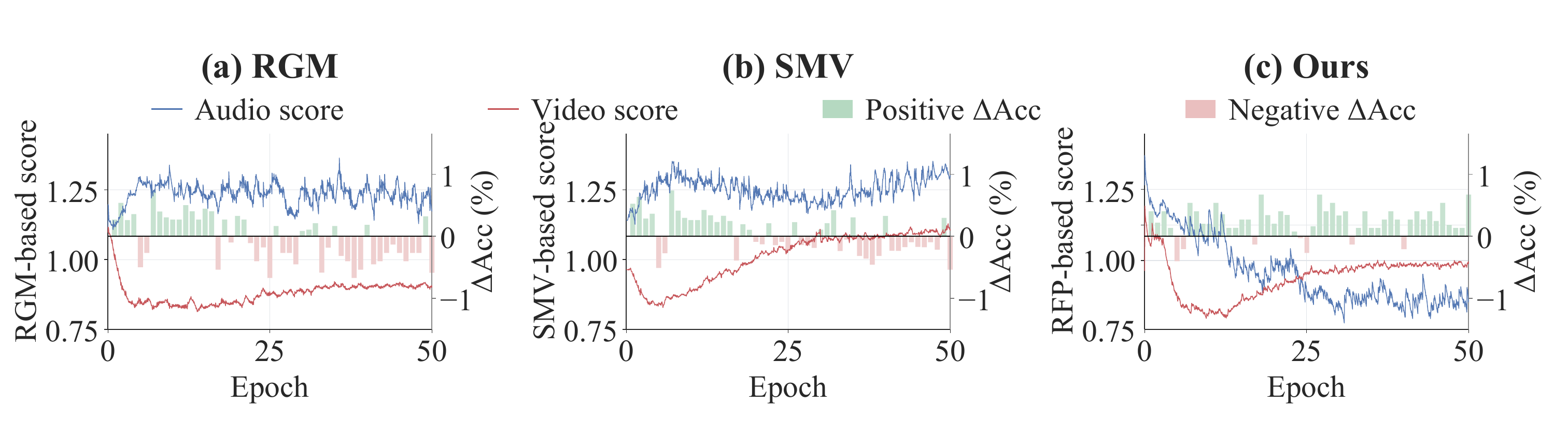}
\caption{
Controlled intervention experiments on CREMA-D. We save 50 checkpoints along a vanilla training trajectory and, from each checkpoint, construct a vanilla continuation and three intervention branches. All branches start from the same model and optimizer states and continue for 10 epochs, with $\Delta$Acc measured relative to the vanilla continuation. Across panels, blue/red curves denote audio/video scores and green/red bars denote positive/negative $\Delta$Acc.
}
\label{fig:empirical_obs}
\end{figure*}

To evaluate the rebalancing reliability of RFP, we conduct controlled intervention experiments on CREMA-D and compare RFP with the modality-level rebalancing scores derived from RGM~\citep{gao2026rgm} and SMV~\citep{wei2024smv}, representing optimization-based and prediction-based signals, respectively. To isolate the quality of these signals, we retain only their modality-level scores and convert all three signals into intervention coefficients using the same mapping rule. Specifically, we select several checkpoints along the same vanilla training trajectory and, from each checkpoint, construct four continuation branches: one vanilla branch that preserves the original optimization process and three rebalancing branches guided by the RGM-derived score, SMV-derived score, and RFP, respectively. We measure the intervention benefit of each signal at the current training stage by comparing the final accuracy of its guided branch with that of the corresponding vanilla continuation. As illustrated in Figure~\ref{fig:empirical_obs}, while all three signals show a consistent audio-over-video ordering with generally beneficial interventions in the early stage of training, the RGM- and SMV-derived scores retain clear audio--video gaps in the later stage even as their interventions become harmful. This suggests that these cross-modal scores can be biased by modality-specific differences and consequently misguide rebalancing. By contrast, the RFP gap gradually narrows and eventually reverses its ordering, while RFP-guided interventions remain beneficial at most checkpoints, suggesting that RFP provides a more comparable measure of optimization progress across modalities.

To this end, we propose \textbf{F}unction-Space \textbf{G}uided \textbf{M}ultimodal \textbf{O}ptimization \textbf{(FGMO)}, which coordinates modality optimization using RFP, a function-space progress signal.
Specifically, we introduce Functional Progress Estimation (FPE) to recover unit-learning-rate updates, measure their tensor-wise responses in function space, and derive RFP. Using the resulting RFP, Functional Response Control (FRC) further allocates modality-level response budgets toward slower-progress modalities under KL regularization. In turn, the assigned budgets are converted into tensor-wise learning rates according to the reference profile and current functional responses, which modulate the optimizer updates. Together, these components form a feedback loop that continuously reassesses and regulates modality progress throughout training.
Our contributions are as follows:
\begin{itemize}
    \item \textbf{We rethink modality imbalance in function space and introduce Relative Functional Progress (RFP)}, which provides a more comparable signal for modality rebalancing.

    \item \textbf{We propose Functional Progress Estimation (FPE) and Functional Response Control (FRC) to form a closed-loop optimization framework,} which dynamically coordinates modality optimization to alleviate modality imbalance.

    \item \textbf{We provide a theoretical analysis
    of the proposed closed-loop controller}
    and establish its one-step target-contraction property for reducing
    RFP discrepancies.
    Extensive experiments on six multimodal benchmarks
    demonstrate the effectiveness of our method.
\end{itemize}

\section{Related Work}

\noindent\textbf{Imbalanced Multimodal Learning.} Compared to unimodal learning, multimodal learning~\citep{baltrusaitis2019multimodal} can integrate multi-source information to improve performance. However, traditional fusion methods ~\citep{baltrusaitis2019multimodal,nie2021multimodal} often suffer from modality imbalance, where the strong modality rapidly dominates the training process, leaving the weak modality under-optimized ~\citep{wang2020gblend,huang2022competition,peng2022ogm}. To alleviate this issue, one line of work estimates modality status from prediction-related signals. OGM~\citep{peng2022ogm} estimates modality imbalance from prediction contributions, while methods such as PMR~\citep{fan2023pmr} and AGM~\citep{li2023agm} introduce modality-specific classifiers or predictions to characterize the learning state of each modality. More recently, ARL~\citep{wei2025arl} and AUG~\citep{jiang2025aug} calibrate modality optimization based on prediction variance and bias and modality classification ability. Another line of work characterizes modality status from optimization-related signals. G-Blend~\citep{wang2020gblend} and Greedy~\citep{wu2022greedy} adjust modality optimization according to generalization dynamics. Recent methods such as AMSS+~\citep{yang2025amss} and RGM~\citep{gao2026rgm} exploit fine-grained parameter information or cumulative gradient energy to regulate modality optimization.
Despite their different forms,
both categories estimate modality imbalance
by directly comparing modality-wise scores.
However, different modalities naturally exhibit
distinct prediction uncertainty and learning dynamics,
making these scores not inherently comparable across modalities.
Consequently, score disparities may reflect
intrinsic modality differences
rather than actual differences in optimization progress.

\par\noindent\textbf{Function-Space Methods.}
To more faithfully characterize how neural networks change during learning,
early studies began to examine optimization from a function-space perspective.
HCGD
compared parameter- and function-space distances during optimization~\citep{benjamin2019function}.
Subsequent works such as S-FSVI
regularized model outputs in function space to preserve previously learned knowledge~\citep{rudner2022sfsvi}.
More recently,
SFR
introduced a function-space parameterization for sequential learning~\citep{scannell2024function},
while FLeRM
proposed an efficient first-order estimator
to quantify function-space changes induced by parameter updates~\citep{milsom2025function}.
Despite these advances,
existing function-space methods mainly focus on single-network or unimodal learning scenarios.
How to leverage function space to measure and balance modality competition
in joint multimodal training remains under-explored.

\section{Method}
\label{sec:method}
\begin{figure*}[t]
\centering
\includegraphics[width=\textwidth]{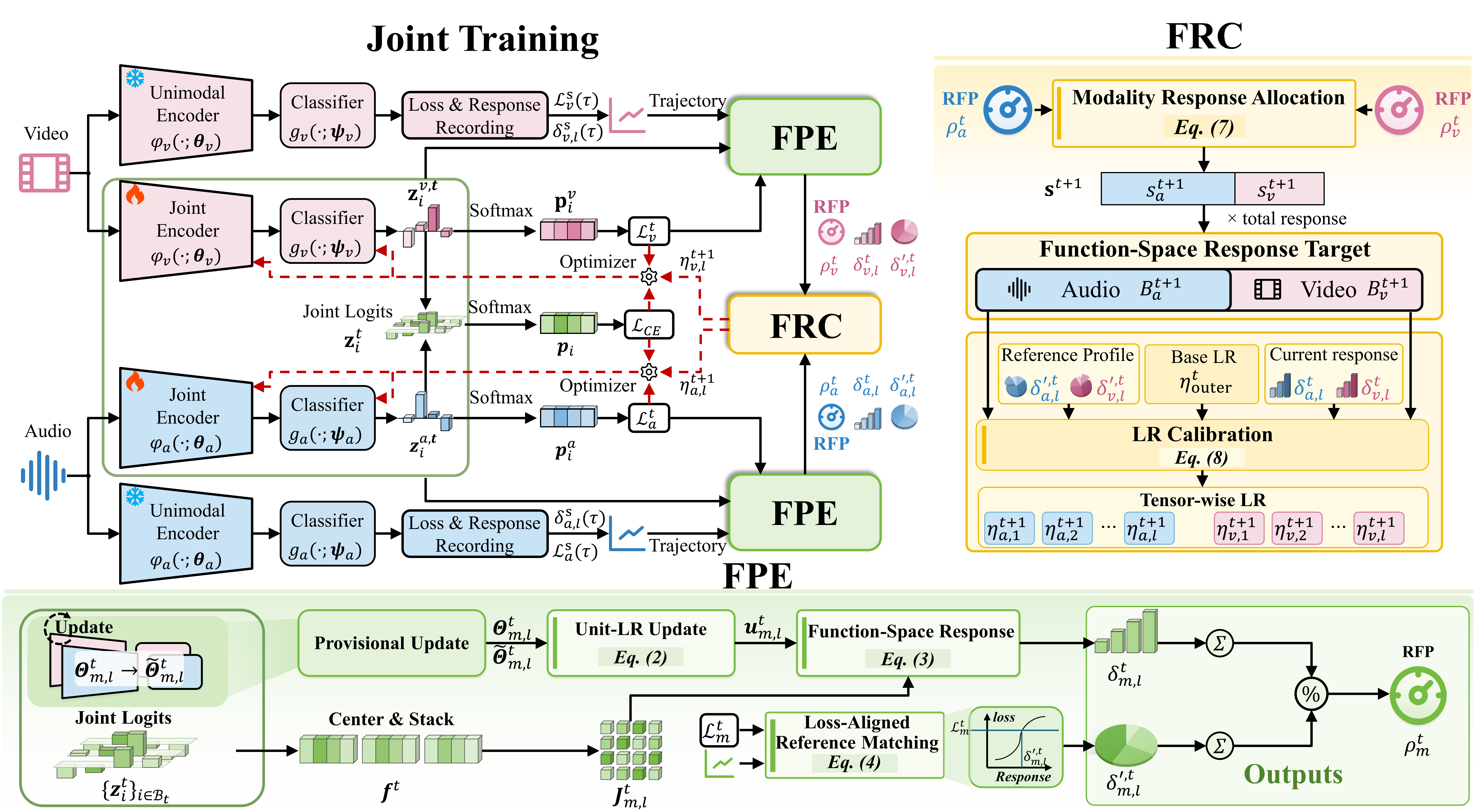}
\caption{The overall framework of Function-Space Guided Multimodal Optimization (FGMO). FPE estimates Relative Functional Progress (RFP) from update-induced function-space responses and loss-aligned unimodal references, while FRC uses RFP feedback to perform modality-level response allocation and tensor-wise learning-rate adjustment.}
\label{fig:pipeline}
\end{figure*}

\TightDisplayMath
In this section, we present Function-Space Guided Multimodal Optimization (FGMO), whose overall framework is illustrated in Figure~\ref{fig:pipeline}. Specifically, we introduce FPE that measures the function-space response and calibrates it against a loss-aligned unimodal reference to obtain RFP. Based on RFP, FRC further redistributes the modality-level function-space response budget and realizes the assigned budget. Together, FPE and FRC form a closed-loop optimization framework that dynamically coordinates modality optimization toward balanced RFP across modalities.

\subsection{Multimodal Learning}
\label{sec:problem_formulation}

We consider a multimodal classification task involving $M$ distinct modalities.
Let $\mathcal{D}=\{(\mathbf{x}_i^{1},\dots,\mathbf{x}_i^{M},\mathbf{y}_i)\}_{i=1}^{N}$ denote a training set of $N$ i.i.d. samples, where $\mathbf{x}_i^{m} \in \mathcal{X}^m$ represents the input of the $m$-th modality and $\mathbf{y}_i \in \{0, 1\}^C$ is the corresponding one-hot label over $C$ classes.
For each modality $m$, a dedicated encoder $\phi_m(\cdot;\boldsymbol{\theta}_m)$ and classifier $g_m(\cdot;\boldsymbol{\psi}_m)$ produce the modality-specific logits
$\mathbf{z}_i^{m}=g_m(\phi_m(\mathbf{x}_i^{m};\boldsymbol{\theta}_m);\boldsymbol{\psi}_m)$,
yielding the unimodal prediction
$\mathbf{p}_i^{m}=\operatorname{softmax}(\mathbf{z}_i^{m})$.
During joint training, the modality-specific logits are averaged to obtain the joint prediction
$\mathbf{p}_i=\operatorname{softmax}\left(\frac{1}{M}\sum_{m=1}^{M}\mathbf{z}_i^{m}\right)$.
Let $\boldsymbol{\Theta} = \{(\boldsymbol{\theta}_m,\boldsymbol{\psi}_m)\}_{m=1}^{M}$ collect all trainable encoder and classifier parameters.
The entire network is optimized with an overall objective combining the multimodal joint loss and the unimodal losses from individual branches:
\begin{equation}
\mathcal{L}(\boldsymbol{\Theta})
=
-\frac{1}{N}
\sum_{i=1}^{N}
\left(
\mathbf{y}_i^\top \log(\mathbf{p}_i)
+
\sum_{m=1}^{M}
\mathbf{y}_i^\top \log(\mathbf{p}_i^{m})
\right).
\label{eq:objective}
\end{equation}

\subsection{Functional Progress Estimation}
\label{sec:fpe}

Based on Eq.~(\ref{eq:objective}), at each optimization update,
the optimizer produces parameter updates from the current mini-batch.
Due to heterogeneous parameterizations and learning dynamics,
these updates cannot directly reflect the optimization progress of each modality.
Hence, FGMO characterizes modality-wise optimization progress
by estimating the function-space response
and calibrating it against a loss-aligned unimodal reference
to obtain the function-space progress signal.

Specifically, we perform progress estimation and response adjustment
every $K$ optimization updates and refer to each such invocation as an adjustment event and index it by $t$.
At adjustment event $t$, let $\mathcal{B}_t$ denote the current mini-batch
and $\mathcal{T}_m$ the set of parameter tensors belonging to modality $m$.
For each $l\in\mathcal{T}_m$,
we first perform a provisional optimizer update on $\mathcal{B}_t$
and then restore the parameters to their pre-update values.
Since we dynamically adjust tensor-wise learning rates during training,
directly using the parameter change after the optimizer update
would make the measured response depend on the current learning-rate scale.
We therefore recover the unit-learning-rate update as:
\begin{equation}
\mathbf{u}_{m,l}^t
=
\frac{
\widetilde{\boldsymbol{\Theta}}_{m,l}^{\,t}
-
\boldsymbol{\Theta}_{m,l}^t
}{
\eta_{m,l}^t
},
\label{eq:unit_update}
\end{equation}
where $\boldsymbol{\Theta}_{m,l}^t$ and $\widetilde{\boldsymbol{\Theta}}_{m,l}^{\,t}$ denote the parameters of the $l$-th tensor for the $m$-th modality before and after the provisional update, $\eta_{m,l}^t$ is the current learning rate for tensor $l$ of modality $m$, and $\mathbf{u}_{m,l}^t$ is the corresponding unit-learning-rate update.

Inspired by an efficient function-space estimator~\citep{milsom2025function},
we characterize the functional effect of each tensor update
through its induced first-order change in the model output.
For the current mini-batch of $n_b$ samples,
let
$\mathbf{z}_i^t=\frac{1}{M}\sum_{m=1}^{M}\mathbf{z}_i^{m,t}$
denote the joint logits of sample $i$, obtained using the same logit-level fusion as in inference.
We subtract the class-wise mean from each logit vector
and stack the resulting centered logits as
$\mathbf{f}^t\in\mathbb{R}^{D}$, where $D=n_bC$.
Let
$\mathbf{J}_{m,l}^t=
\partial\mathbf{f}^t/\partial\boldsymbol{\Theta}_{m,l}^t$
denote the Jacobian with respect to tensor $l$ of modality $m$.
Under a first-order approximation, we define
the tensor-wise function-space response as:
\begin{equation}
\delta_{m,l}^t
=
\frac{1}{\sqrt{D}}
\left\|
\mathbf{J}_{m,l}^t\mathbf{u}_{m,l}^t
\right\|_2 ,
\label{eq:function_response}
\end{equation}
which measures the normalized functional change induced by
the unit-learning-rate update.
In practice, we avoid explicitly constructing
$\mathbf{J}_{m,l}^t$ and estimate $\delta_{m,l}^t$
using Gaussian random projection with a low-variance
Kronecker-factored EMA estimator~\citep{milsom2025function}.

To construct the unimodal reference,
we independently train each modality and evaluate
the unimodal loss $\mathcal{L}_m^{\mathrm{s}}(\tau)$
and tensor-wise response $\delta_{m,l}^{\mathrm{s}}(\tau)$
every $K$ optimization updates,
where $\tau$ indexes the corresponding response estimation event.
Meanwhile, during joint training, we align the joint and unimodal optimization states by loss rather than epoch:
\begin{equation}
\tau_m^t
=
\arg\min_{\tau}
\left|
\log\mathcal{L}_m^t
-
\log\mathcal{L}_m^{\mathrm{s}}(\tau)
\right|,
\label{eq:tau}
\end{equation}
where $\mathcal{L}_m^t$ denotes the modality-specific loss during joint training,
and $\delta_{m,l}^{\prime,t}=\delta_{m,l}^{\mathrm{s}}(\tau_m^t)$ denotes the matched unimodal response.
Using the current and reference responses,
we obtain the Relative Functional Progress (RFP), denoted by $\rho_m^t$,
of modality $m$ as:
\begin{equation}
\rho_m^t
=
\frac{
\sum_{l\in\mathcal{T}_m}\delta_{m,l}^t+\epsilon
}{
\sum_{l\in\mathcal{T}_m}\delta_{m,l}^{\prime,t}+\epsilon
},
\label{eq:rfp}
\end{equation}
where $\epsilon>0$ is a small constant for numerical stability.
The ratio measures the overall function-space response during joint training
relative to the modality's own reference response,
reducing the influence of modality-specific response scales
and enabling comparison through relative rather than absolute progress.
The resulting RFP is then passed to FRC
as the feedback signal for subsequent function-space response allocation.

\subsection{Functional Response Control}
\label{sec:frc}

Through Eq.~(\ref{eq:rfp}) in FPE,
we obtain the RFP $\rho_m^t$
of each modality at the current optimization state.
To use this feedback to coordinate subsequent optimization,
we propose Functional Response Control (FRC),
which dynamically redistributes modality-level function-space responses according to RFP
and further realizes the assigned response targets
through tensor-wise learning-rate adjustment.

Specifically, at adjustment event $t$,
we use $\mathbf{s}^t=(s_1^t,\ldots,s_M^t)\in\Delta_M$
to represent the current modality response shares,
where $s_m^t$ denotes the response fraction assigned to modality $m$.
Since a smaller RFP indicates slower functional progress
relative to its unimodal reference,
directly favoring modalities with smaller RFP
may over-concentrate the response
and ignore the previous allocation state.
We therefore use $\log\rho_m^t$ as the progress cost
and constrain the change from the previous allocation
using the Kullback--Leibler divergence:
\begin{equation}
\mathbf{s}^{t+1}=\arg\min_{\mathbf{s}\in\Delta_M}\left[\sum_{m=1}^{M}s_m\log\rho_m^t+\frac{1}{\lambda}D_{\mathrm{KL}}(\mathbf{s}\|\mathbf{s}^t)\right].
\label{eq:kl_objective}
\end{equation}
where $\lambda>0$ controls the response strength of FRC
to the current RFP.
The first term adjusts modality responses
according to the current relative functional progress,
while the KL term preserves the allocation state
across consecutive adjustment events.
The objective admits the closed-form solution:
\begin{equation}
s_m^{t+1}=\frac{s_m^t(\rho_m^t)^{-\lambda}}{\sum_{j=1}^{M}s_j^t(\rho_j^t)^{-\lambda}}.
\label{eq:share_update}
\end{equation}
This update adapts the modality response allocation
to the current RFP as relative progress evolves during training.
We then convert the updated response share
into the modality-level function-space response target
$B_m^{t+1}=s_m^{t+1}\sum_{k=1}^{M}\sum_{l\in\mathcal{T}_k}\delta_{k,l}^t$,
where $B_m^{t+1}$ denotes the function-space response
assigned to modality $m$ at the next adjustment.

Accordingly, to realize the modality-level response target
across parameter tensors,
we use the loss-aligned unimodal response matched by Eq.~(\ref{eq:tau})
to define the loss-aligned reference profile of modality $m$, whose entry for tensor $l$ is
$\delta_{m,l}^{\prime,t}/\sum_{j\in\mathcal{T}_m}\delta_{m,j}^{\prime,t}$.
Since $\delta_{m,l}^t$ characterizes the function-space response
under a unit learning rate,
we scale the learning rate inversely with the current response
to realize the assigned tensor-level target:
\begin{equation}
\eta_{m,l}^{t+1}=\eta_{\mathrm{outer}}^t\frac{B_m^{t+1}\delta_{m,l}^{\prime,t}}{(\delta_{m,l}^t+\epsilon)\sum_{j\in\mathcal{T}_m}\delta_{m,j}^{\prime,t}},\qquad l\in\mathcal{T}_m,
\label{eq:final_lr}
\end{equation}
where $\eta_{\mathrm{outer}}^t$ denotes the base learning rate.
Finally, the tensor-wise learning rates produced by FRC rescale the parameter updates generated by the optimizer, allowing each tensor to be updated along the optimizer direction with its assigned step size.
In this way,
FRC maps the modality-level response target
to tensor-wise learning rates
while accounting for heterogeneous functional responses across tensors.

\subsection{Theoretical Analysis}
\label{sec:theory}

We analyze FRC via one-step target contraction to characterize whether its response target reduces the current modality optimization discrepancy. For clarity, we consider two modalities with input RFP ratio $Q^t=\rho_1^t/\rho_2^t$, where $\log Q^t=0$ indicates balanced relative functional progress, and let $Q_{\mathrm{tar}}^{t+1|t}$ denote the FRC-assigned target ratio.

\textbf{Theorem 1 (One-Step Target Contraction, Informal).}
The target ratio satisfies
\begin{equation}
\log Q_{\mathrm{tar}}^{t+1|t}
=
(1-\lambda)\log Q^t+d_t,
\label{eq:rfp_convergence}
\end{equation}
where $\lambda$ is the controller gain and $d_t$ is the controller-state mismatch. If $0<\lambda<2$ and $|d_t|<(1-|1-\lambda|)|\log Q^t|$, then
\begin{equation}
|\log Q_{\mathrm{tar}}^{t+1|t}|
<
|\log Q^t|.
\end{equation}

Theorem~1 shows that, under bounded controller-state mismatch and $0<\lambda<2$, the FRC target strictly contracts toward balance. This theoretically characterizes FRC as a negative-feedback controller over RFP, supporting the use of RFP feedback to coordinate modality optimization.

\section{Experiments}
\NormalDisplayMath

\subsection{Experimental Setup}

\textbf{Dataset:}
We conduct experiments on six widely used multimodal datasets, i.e., CREMA-D~\citep{cao2014cremad}, KSounds~\citep{arandjelovic2017look}, VGGSound~\citep{chen2020vggsound}, Twitter~\citep{yu2019twitter}, Sarcasm~\citep{cai2019sarcasm}, and NVGesture~\citep{molchanov2016nvgesture}, covering audio-video, image-text, and trimodal settings. CREMA-D is an audio-visual dataset for emotion recognition, containing 7,442 video clips from 91 actors. KSounds is an audio-visual dataset for human action recognition with 31 action classes. VGGSound is a large-scale audio-visual dataset with nearly 200K video clips. Twitter is an image-text dataset for emotion recognition with three classes. Sarcasm is designed for sarcasm detection and contains 24,635 image-text pairs. NVGesture is a trimodal dataset for gesture recognition, containing 1,532 samples from RGB, optical flow (OF), and depth modalities.

\textbf{Baselines:}
We compare FGMO with traditional multimodal fusion methods, including Concat~\citep{baltrusaitis2019multimodal}, Affine~\citep{perez2018film}, and ML-LSTM~\citep{nie2021multimodal}, and modality rebalancing methods, including G-Blend~\citep{wang2020gblend}, MSLR~\citep{yao2022mslr}, OGM~\citep{peng2022ogm}, PMR~\citep{fan2023pmr}, AGM~\citep{li2023agm}, MMPareto~\citep{wei2024mmpareto}, 
SMV~\citep{wei2024smv},
MLA~\citep{zhang2024mla},
DI-MML~\citep{fan2024dimml}, ReconBoost~\citep{hua2024reconboost},
LFM~\citep{yang2024lfm}, AMSS+~\citep{yang2025amss}, InfoReg~\citep{huang2025inforeg}, ARL~\citep{wei2025arl}, AUG~\citep{jiang2025aug}, 
DecAlign~\citep{qian2026decalign}
and RGM~\citep{gao2026rgm}.

\textbf{Evaluation Protocols:}
Following established evaluation protocols in prior multimodal studies~\citep{yu2019twitter,cai2019sarcasm,peng2022ogm}, we adopt accuracy, mean average precision (mAP), and Macro-F1 as evaluation metrics. Accuracy measures the proportion of correctly classified samples. mAP reports the mean average precision over all classes, while Macro-F1 computes the average F1 score across classes.

\textbf{Implementation Details:}
Following OGM~\citep{peng2022ogm}, we employ ResNet-18~\citep{he2016resnet} as the backbone for both audio and visual encoders on CREMA-D, KSounds, and VGGSound. All backbone parameters are randomly initialized. For NVGesture, we employ I3D~\citep{carreira2017i3d} as the encoder for RGB, OF, and depth modalities and initialize it with pretrained weights. We use SGD with a momentum of $0.9$ and a weight decay of $1\times10^{-4}$ for these datasets. The initial learning rates are $7.5\times10^{-3}$ for CREMA-D, $1\times10^{-2}$ for KSounds and VGGSound, and $8\times10^{-3}$ for NVGesture. For Twitter and Sarcasm, we adopt BERT~\citep{devlin2019bert} as the text encoder and ResNet-50~\citep{he2016resnet} as the image encoder. We use Adam~\citep{kingma2015adam} with an initial learning rate of $2\times10^{-5}$. The batch sizes are set to 64, 32, 16, 2, 32, and 64 for CREMA-D, KSounds, VGGSound, NVGesture, Twitter, and Sarcasm, respectively. For fair comparison, all baselines use the same backbone and initialization strategy. All models are implemented with PyTorch~\citep{paszke2019pytorch} and experiments are conducted on NVIDIA GeForce RTX 4090 GPUs.

\begin{table*}[t]
\centering
\caption{Comparison with SOTA multimodal learning methods. The best results are highlighted in \textbf{bold}. The \underline{underline} denotes the second-best performance. Gray-shaded results indicate that multimodal performance is lower than that of the best unimodal approach.}
\label{tab:main_results}
\scriptsize
\setlength{\tabcolsep}{3.5pt}
\renewcommand{\arraystretch}{1.05}
\resizebox{\textwidth}{!}{
\begin{tabular}{l|cc|cc|cc|cc|cc|cc}
\toprule
\multirow{2}{*}{\textbf{Method}} 
& \multicolumn{2}{c|}{CREMA-D} 
& \multicolumn{2}{c|}{KSounds} 
& \multicolumn{2}{c|}{VGGSound} 
& \multicolumn{2}{c|}{Twitter} 
& \multicolumn{2}{c|}{Sarcasm} 
& \multicolumn{2}{c}{NVGesture} \\
\cline{2-13}
& Acc. & mAP 
& Acc. & mAP 
& Acc. & mAP 
& Acc. & F1 
& Acc. & F1 
& Acc. & F1 \\
\midrule

Unimodal-1 
& .6317 & .5879 
& .5412 & .5669 
& .4655 & .4701 
& .5863 & .4333 
& .7181 & .7073 
& .7822 & .7833 \\

Unimodal-2 
& .4583 & .6861 
& .5562 & .5837 
& .3494 & .3478 
& .7367 & .6849 
& .8136 & .8056 
& .7863 & .7865 \\

Unimodal-3 
& -- & -- 
& -- & -- 
& -- & -- 
& -- & -- 
& -- & -- 
& .8154 & .8183 \\

\midrule

Concat 
& .6361 & \cellcolor{gray!20}.6841 
& .6455 & .7130 
& .5116 & .5352 
& \cellcolor{gray!20}.7011 & \cellcolor{gray!20}.6386 
& .8286 & .8240 
& .8237 & .8270 \\

Affine 
& .6626 & .7193 
& .6424 & .6931 
& .5001 & .5155 
& \cellcolor{gray!20}.7203 & \cellcolor{gray!20}.5992 
& .8240 & .8188 
& .8278 & .8281 \\

ML-LSTM 
& \cellcolor{gray!20}.6290 & \cellcolor{gray!20}.6473 
& .6394 & .6902 
& .4966 & .5139 
& \cellcolor{gray!20}.7068 & \cellcolor{gray!20}.6564 
& .8277 & .8205 
& .8320 & .8330 \\

\midrule

G-Blend 
& .6465 & .7392 
& .6722 & .7274 
& .5086 & .5555 
& \cellcolor{gray!20}.7309 & \cellcolor{gray!20}.6799 
& .8286 & .8215 
& .8299 & .8305 \\

MSLR 
& .6868 & .7412 
& .6756 & .7282 
& .4987 & .5415 
& \cellcolor{gray!20}.7232 & \cellcolor{gray!20}.6382 
& .8439 & .8378 
& .8237 & .8284 \\

OGM 
& .6612 & .7372 
& .6582 & .7159 
& .4829 & .4978 
& \cellcolor{gray!20}.7058 & \cellcolor{gray!20}.6435 
& .8360 & .8293 
& -- & -- \\

PMR 
& .6659 & .7058 
& .6675 & .7274 
& \cellcolor{gray!20}.4647 & .4866 
& \cellcolor{gray!20}.7357 & \cellcolor{gray!20}.6636 
& .8310 & .8256 
& -- & -- \\

AGM 
& .6733 & .7807 
& .6791 & .7388 
& .4711 & .5198 
& \cellcolor{gray!20}.7261 & \cellcolor{gray!20}.6502 
& .8306 & .8293 
& .8279 & .8284 \\

MMPareto 
& .7487 & .8535 
& .7000 & .7850 
& .5125 & .5473 
& \cellcolor{gray!20}.7358 & \cellcolor{gray!20}.6729 
& .8348 & .8284 
& .8382 & .8424 \\

SMV 
& .7872 & .8417 
& .6900 & .7426 
& .5031 & .5362 
& .7428 & \cellcolor{gray!20}.6817 
& .8418 & .8368 
& .8352 & .8341 \\

MLA 
& .7943 & .8572 
& .7004 & .7945 
& .5165 & .5473 
& \cellcolor{gray!20}.7352 & \cellcolor{gray!20}.6713 
& .8426 & .8348 
& .8340 & .8372 \\

DI-MML 
& .8158 & .8592 
& .7203 & .7426 
& .5173 & .5479 
& \cellcolor{gray!20}.7248 & \cellcolor{gray!20}.6686 
& .8411 & .8315 
& -- & -- \\

ReconBoost 
& .7557 & .8140 
& .6855 & .7662 
& .5097 & .5387 
& .7442 & \cellcolor{gray!20}.6832 
& .8437 & .8317 
& .8386 & .8434 \\

LFM 
& .8362 & .9006 
& .7253 & .7897 
& .5274 & .5694 
& .7501 & \underline{.7057} 
& .8497 & .8457 
& .8436 & .8468 \\

AMSS+ 
& .7030 & .7641 
& .7255 & .7913 
& -- & -- 
& \underline{.7558} & .6981 
& .8435 & .8377 
& .8464 & .8494 \\

InfoReg 
& .7498 & .8603 
& .7189 & .7986 
& .5289 & .5712 
& .7403 & \cellcolor{gray!20}.6847 
& .8479 & .8396 
& -- & -- \\

ARL 
& .8379 & .8927 
& .7198 & .7826 
& .5234 & .5583 
& .7386 & \cellcolor{gray!20}.6786 
& -- & -- 
& .8375 & .8432 \\

AUG 
& \underline{.8515} & \underline{.9103} 
& .7263 & .7901 
& \underline{.5301} & \underline{.5826} 
& .7512 & .6962 
& \underline{.8510} & \underline{.8458} 
& \underline{.8501} & \underline{.8533} \\

DecAlign
& .8426 & .9027 
& .7298 & .7926 
& .5268 & .5677 
& .7396 & \cellcolor{gray!20}.6786 
& -- & -- 
& .8397 & .8429 \\

RGM 
& .8327 & .8993 
& \underline{.7309} & \underline{.8088} 
& -- & -- 
& .7512 & .7035 
& .8449 & .8397 
& .8406 & .8470 \\

\midrule

\multirow{2}{*}{\textbf{FGMO}} 
& \textbf{.8589} & \textbf{.9126} 
& \textbf{.7565} & \textbf{.8210} 
& \textbf{.5341} & \textbf{.5864} 
& \textbf{.7628} & \textbf{.7131} 
& \textbf{.8547} & \textbf{.8486} 
& \textbf{.8548} & \textbf{.8582} \\[-0.8ex]

& {\tiny $\pm0.0037$} 
& {\tiny $\pm0.0012$} 
& {\tiny $\pm0.0033$} 
& {\tiny $\pm0.0058$} 
& {\tiny $\pm0.0023$} 
& {\tiny $\pm0.0024$} 
& {\tiny $\pm0.0046$} 
& {\tiny $\pm0.0042$} 
& {\tiny $\pm0.0021$} 
& {\tiny $\pm0.0012$} 
& {\tiny $\pm0.0034$} 
& {\tiny $\pm0.0028$} \\

\bottomrule
\end{tabular}
}
\end{table*}

\subsection{Main Results}

\textbf{Comparison with Multimodal Baselines:}
Table~\ref{tab:main_results} reports the results on all six datasets. Unimodal-1/2 denote audio/video for CREMA-D, KSounds, and VGGSound, and image/text for Twitter and Sarcasm, respectively. Unimodal-1/2/3 denote RGB/OF/Depth for NVGesture, respectively. Gray-shaded entries indicate multimodal results below the best unimodal performance on the corresponding metric.
Simple fusion methods, including Concat~\citep{baltrusaitis2019multimodal} and Affine~\citep{perez2018film}, underperform the best unimodal model on Twitter, showing that combining modalities alone does not guarantee better performance.
Existing modality rebalancing methods improve performance in many settings, but their relative performance varies across datasets and metrics.
In contrast, \method\ achieves the best results on both reported metrics across all six datasets, outperforming competitive methods such as AUG~\citep{jiang2025aug} and RGM~\citep{gao2026rgm}.
Notably, on KSounds, \method\ achieves an accuracy of 0.7565 and an mAP of 0.8210, exceeding the strongest baseline, RGM, by 2.56\% and 1.22\%, respectively.
These results support the effectiveness of our reference-calibrated function-space rebalancing across different modality combinations.

\subsection{Ablation Study}

We conduct ablation studies on Sarcasm and KSounds to evaluate four key designs in FGMO:
function-space response measurement, the relative progress metric RFP,
loss-aligned reference matching, and tensor-wise response allocation.
The results are reported in Table~\ref{tab:ablation}.

\begin{table*}[t]
\centering
\scriptsize
\setlength{\tabcolsep}{3pt}
\renewcommand{\arraystretch}{1.05}

\begin{minipage}[t]{0.73\linewidth}
\vspace{0pt} 
\caption{Ablation study on Sarcasm and KSounds. We ablate function-space measurement, relative progress normalization (RFP), loss-aligned reference matching, and tensor-wise allocation.}
\label{tab:ablation}
\end{minipage}\hfill
\begin{minipage}[t]{0.25\linewidth}
\vspace{0pt} 
\captionof{figure}{Training time and accuracy under different $K$ on CREMA-D dataset.}
\label{fig:training_time_k}
\end{minipage}

\vspace{-12pt}

\begin{minipage}[t]{0.73\linewidth}
\vspace{0pt} 
\centering
\resizebox{\linewidth}{!}{
\begin{tabular}{lcccc|ccc|ccc}
\toprule
\multirow{2}{*}{\textbf{Variant}} & \multirow{2}{*}{\textbf{Space}} & \multirow{2}{*}{\textbf{Align.}} & \multirow{2}{*}{\textbf{Metric}} & \multirow{2}{*}{\textbf{Granularity}} & \multicolumn{3}{c|}{\textbf{Sarcasm (Acc.)}} & \multicolumn{3}{c}{\textbf{KSounds (Acc.)}} \\
\cmidrule(lr){6-8} \cmidrule(lr){9-11}
& & & & & \textbf{Multi} & \textbf{Image} & \textbf{Text} & \textbf{Multi} & \textbf{Audio} & \textbf{Video} \\
\midrule
Baseline & -- & -- & -- & -- & 0.8286 & 0.6907 & 0.8066 & 0.6455 & 0.4998 & 0.4798 \\
(a) w/o Function-space & Param. & Loss & Relative & Tensor & 0.8355 & 0.7001 & 0.8120 & 0.6790 & 0.5207 & 0.5490 \\
(b) w/o RFP & Func. & Loss & Absolute & Tensor & 0.8348 & 0.6978 & 0.8132 & 0.6952 & 0.5191 & 0.5508 \\
(c) w/o Loss-Align & Func. & Epoch & Relative & Tensor & 0.8389 & 0.7024 & 0.8177 & 0.7352 & 0.5276 & 0.5551 \\
(d) w/o Tensor-wise & Func. & Loss & Relative & Modality & 0.8427 & 0.7082 & 0.8223 & 0.7449 & 0.5222 & 0.5613 \\
\midrule
\rowcolor{gray!15} \textbf{FGMO (Ours)} & \textbf{Func.} & \textbf{Loss} & \textbf{Relative} & \textbf{Tensor} & \textbf{0.8547} & \textbf{0.7165} & \textbf{0.8281} & \textbf{0.7565} & \textbf{0.5319} & \textbf{0.5991} \\
\bottomrule
\end{tabular}
}
\end{minipage}\hfill
\begin{minipage}[t]{0.25\linewidth}
\vspace{0pt} 
\centering
\includegraphics[width=\linewidth]{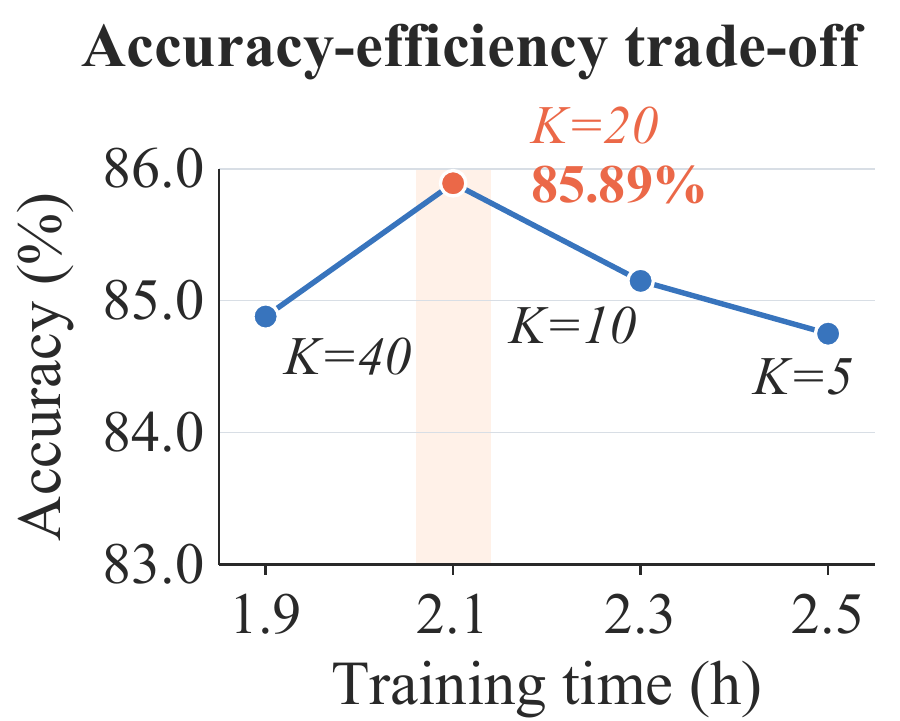}
\end{minipage}

\end{table*}

As shown in Table~\ref{tab:ablation}, removing any of the four designs consistently reduces multimodal performance. Replacing function-space responses with parameter-space update magnitudes reduces multimodal accuracy by 1.92\% and 7.75\% on Sarcasm and KSounds, respectively, supporting the benefit of function-space measurement even when reference normalization is retained. Replacing RFP with absolute function-space responses reduces multimodal accuracy by 1.99\% and 6.13\% on Sarcasm and KSounds, respectively, which highlights the importance of measuring modality progress relative to its own reference rather than directly comparing raw response magnitudes. Replacing loss alignment with epoch alignment also decreases performance, indicating that matching comparable optimization states is important for reference calibration. Finally, removing tensor-wise allocation consistently lowers accuracy, showing that fine-grained response control provides additional gains beyond modality-level allocation. Combining all four designs achieves the best multimodal and modality-specific performance on both datasets.

\subsection{Hyperparameter Analysis}

We investigate the influence of the adjustment interval $K$ and the controller gain $\lambda$ on the model performance and its internal optimization dynamics on the CREMA-D and Sarcasm datasets.

\textbf{Analysis of Adjustment Interval $K$:}
The interval $K$ controls how frequently FGMO estimates RFP and invokes FRC to update the modality budgets. A small $K$ provides frequent feedback, but makes the controller more responsive to short-term optimization fluctuations and increases the function-space probing overhead. In contrast, a large $K$ reduces the computational cost but introduces delayed feedback, limiting timely correction of evolving modality imbalance. As shown in Figure~\ref{fig:training_time_k}, $K=20$ achieves the highest accuracy of 0.8589 with a training time of 2.1796 hours. These results suggest that a moderate update frequency provides a better balance among feedback stability, control responsiveness, and training efficiency. We therefore set $K=20$ as the default.

\begin{figure}[htbp]
\centering
\includegraphics[width=1.0\linewidth]{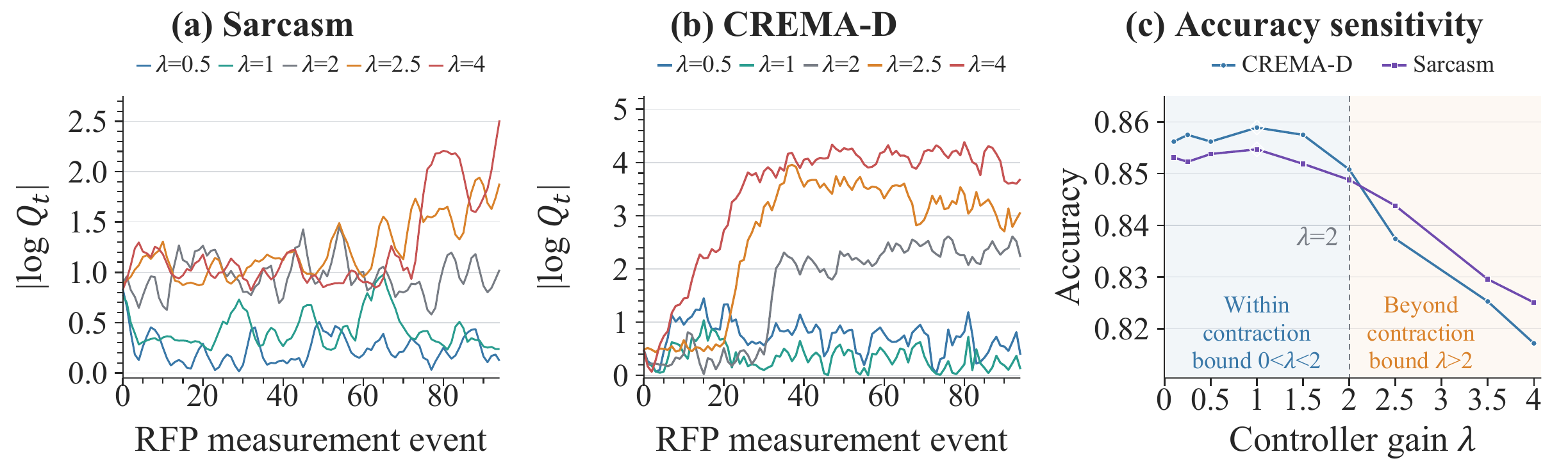}
\caption{Sensitivity analysis of the controller gain $\lambda$ on the Sarcasm and CREMA-D datasets. (a) and (b) Evolution of the absolute log RFP ratio $|\log Q^t|$ under different controller gains on Sarcasm and CREMA-D, respectively. (c) Impact of $\lambda$ on the multimodal classification accuracy. The dashed line ($\lambda=2$) denotes the boundary of the
ideal target-contraction range.}
\label{fig:sensitivity}
\end{figure}

\textbf{Analysis of Controller Gain $\lambda$:}
To empirically examine the target-contraction behavior characterized in
\textbf{Theorem 1}, we evaluate the controller gain across a broad range
$\lambda \in \{0.1, 0.25, 0.5, 1, 1.5, 2, 2.5, 3.5, 4\}$.
As illustrated in Figure~\ref{fig:sensitivity} (c), when $\lambda$ falls
within the theoretically derived target-contraction range
($0 < \lambda < 2$), the model accuracy remains stably high, peaking at
$\lambda=1$ (0.8589, 0.8547).
However, once $\lambda$ reaches or exceeds this range ($\lambda \ge 2$),
the performance degrades significantly.
Figure~\ref{fig:sensitivity} (a), (b) further shows training dynamics
consistent with the theoretical analysis.
Within the target-contraction range (e.g., $\lambda=0.5,1$), the
closed-loop controller effectively maintains substantially smaller RFP discrepancies.
Conversely, an excessively large gain (e.g., $\lambda=2.5,4$) lies outside
the target-contraction range and is accompanied by pronounced oscillations
and an enlarged RFP discrepancy.

\subsection{Further Analysis}

\begin{figure*}[t]
\centering
\includegraphics[width=\textwidth]{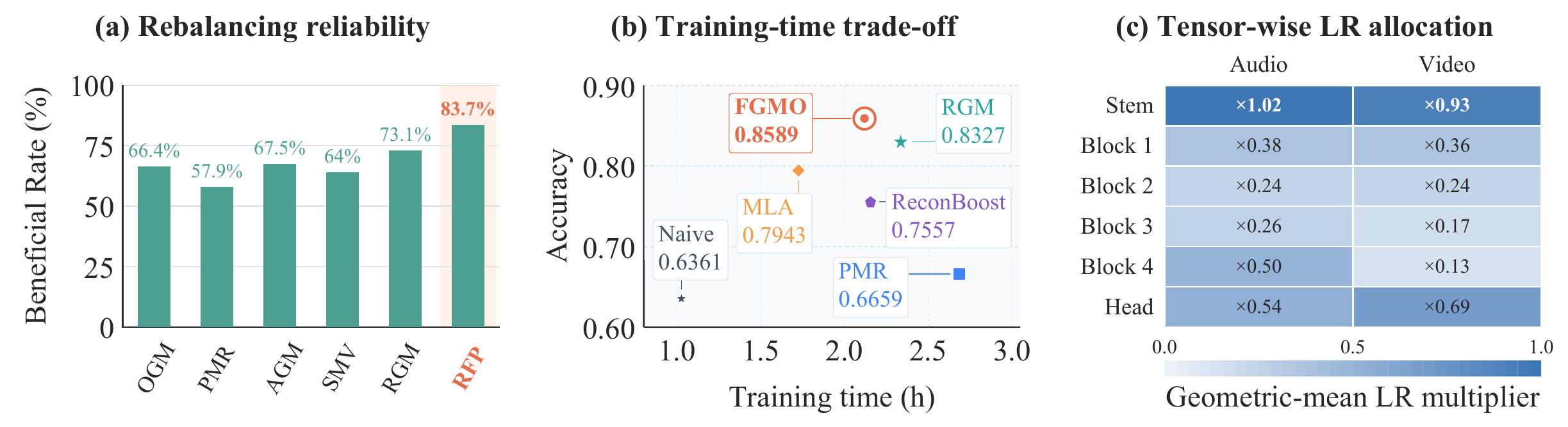}
\caption{
(a) Beneficial Rates of different modality signals under controlled interventions.
(b) Training-time comparison with representative multimodal learning methods.
(c) Tensor-wise learning-rate allocation across the audio and video branches, where each cell reports the geometric-mean learning-rate multiplier.
}
\label{fig:further_analysis}
\end{figure*}

\textbf{Reliability of Rebalancing Signals.} To evaluate signal reliability across training stages, we extend the controlled interventions (Figure~\ref{fig:empirical_obs}) to six datasets and $1{,}000$ checkpoints. We compare RFP with five baselines using the \emph{Beneficial Rate}---the percentage of checkpoints where signal-guided interventions improve accuracy over vanilla continuations. As Figure~\ref{fig:further_analysis} (a) shows, RFP achieves the highest beneficial rate of $83.7\%$, outperforming the competitive baselines and demonstrating superior and more reliable guidance for stage-dependent modality rebalancing.

\textbf{Computational Cost.}
FGMO introduces additional computation for unimodal reference construction and periodic function-space probing.
Figure~\ref{fig:further_analysis} (b) compares joint-training time, excluding reference construction.
The same fixed references are reused across training seeds, allowing their construction cost to be amortized over repeated runs under the same setting.
Although reference construction increases the end-to-end training cost, FGMO achieves improved multimodal performance through reference-calibrated rebalancing.

\textbf{Interpretable LR Allocation.} As Figure~\ref{fig:further_analysis} (c) shows, FRC exhibits a clear hierarchical pattern in learning-rate allocation rather than a simple depth-wise decay. Stem layers retain multipliers near $1\times$, suggesting FRC preserves the adaptation capacity of low-level feature extractors. Meanwhile, intermediate and deep blocks are substantially suppressed to stabilize high-level representations, before the multipliers increase again at the prediction heads. Moreover, audio and video show distinct allocation patterns in deeper blocks, confirming that FRC dynamically learns a modality-specific optimization strategy instead of applying a predefined schedule.

\section{Conclusion}
In this paper, we propose Function-Space Guided Multimodal Optimization (FGMO) to alleviate modality imbalance through comparable progress estimation and coordinated optimization. Functional Progress Estimation (FPE) measures update-induced function-space responses and calibrates them against loss-aligned unimodal references to produce a comparable progress signal termed RFP. Guided by this signal, Functional Response Control (FRC) redistributes modality-level function-space budgets and realizes target responses through tensor-wise learning-rate adjustment. Theoretical analysis establishes a one-step target-contraction property of FRC under bounded controller-state mismatch, and extensive experiments demonstrate the effectiveness of FGMO across multiple multimodal benchmarks.

\section*{AI Use Statement}

We used GPT solely to assist with language polishing and improving the readability of the manuscript. All AI-assisted edits were reviewed and verified by the authors, who take full responsibility for the final content of this work.

\bibliography{references}
\bibliographystyle{plainnat}

\end{document}